# Trainable Reference-Based Evaluation Metric for Identifying Quality of English-Gujarati Machine Translation System

Nisheeth Joshi[1,2,a], Pragya Katyayan[1,2,b], Palak Arora[1,2,c]

Author Affiliations
[1]*Speech and Language Processing Lab, Centre for Artificial Intelligence, Banasthali Vidyapith, Rajasthan.*
[2]*Department of Computer Science, Banasthali Vidyapith, Rajasthan.*

*Author Emails*
[a]*Corresponding author: nisheeth.joshi@rediffmail.com*
[b]*pragya.katyayan@outlook.com*
[c]*palak.arora.pa55@gmial.com*

## Abstract

Machine Translation (MT) Evaluation is an integral part of the MT development life cycle. Without analyzing the outputs of MT engines, it is impossible to performance of an MT system. Through experiments, it has been identified that what works for English and other European languages does not work well with Indian languages. Thus, In this paper, we have introduced a reference-based MT evaluation metric for Gujarati which is based on supervised learning. We have trained two versions of the metric which uses 25 features for training. Among the two models, one model is trained using 6 hidden layers with 500 epochs while the other model is trained using 10 hidden layers with 500 epochs. To test the performance of the metric, we collected 1000 MT outputs of seven MT systems. These MT engine outputs were compared with 1 human reference translation. While comparing the developed metrics with other available metrics, it was found that the metrics produced better human correlations.

## Introduction

The Evaluation of MT systems is an imperative field of research for optimizing the performance of MT systems. It's an evaluation method that helps us determine the effectiveness of MT systems. There are two primary approaches for evaluating such systems: human evaluation and automatic evaluation. Human evaluation is considered the best way to appraise translation quality, but it is time-consuming and costly. Moreover, it gets difficult to find reliable annotators who can precisely evaluate the MT system outputs as the process is very tiring and requires a lot of manual human effort. On the other hand, automatic evaluation is comparatively fast, cost-effective, and easy to use, but at the same time suffers from quality issues. In literature, one can find a variety of MT evaluation metrics viz BLEU, NIST, TER, METEOR, BERTScore, BLEURT, BEER, COMET etc. All these metrics suffer from some or the other issues. The goal of research in MT evaluation is to develop algorithms (metrics) or machine learning models that can produce results at par with human evaluation.

In general, all automatic MT evaluation metrics compare the MT system's output with human-generated reference translations. Here, a test dataset is developed where source language sentences are taken, and some human annotators are asked to provide manual translations for these in the target language. Next, these source sentences are then fed to MT systems and their outputs are registered. Once we have human reference translations and MT outputs from several MT systems, we can compare their performance using one or more automatic MT evaluation metrics.

In this research, we have introduced a reference-based MT evaluation metric that uses the power of deep learning and is named MATRA (MAchine TRanslation Analysis), specially designed for Gujarati. Two variants of this metric were trained by utilizing 25 features.

## Literature Review

Papineni et al. (2002) from IBM Research introduced BLEU (BiLingual Evaluation Understudy) metric for automatic evaluation of Machine Trasnlation outputs [2]. They highlighted that human evaluation takes a lot of time and is expensive. They proposed an automatic machine translation evaluation metric which is fast, cheap and independent of any language. Its strength was its ability to correlate with human judgement at a high degree. It considered modified n-grams precision on blocks of text, sentence length and sentence brevity penalty. Lin and Och (2004) introduced ORANGE as a new automatic evaluation metric for MT outputs [3]. This new metric used no human intervention, was applicable on sentence-level, used many existing data-points, and had the objective to optimize which made it a natural fit for SMT framework. Banerjee and Lavie (2005) introduced METEOR as an automatic MT evaluation metric based on unigram matching concept [4]. It considered their surface and stemmed forms along with meaning of the sentences. METEOR was also extendable to include advanced matching strategies.

Gupta et al. (2010) presented METEOR-Hindi with Hindi as a target language [5]. They changed METEOR's alignment algorithm and scoring mechanism to better suit the needs of Hindi language. This metric showed better correlation with human judgement than BLEU. Joshi et al. (2013) proposed a new human evaluation metric-HEval to address issues like essential requirement of reference translation to compare MT outputs with [6]. HEval metric provided human experts with Heval metric provided human experts with solid grounds to make sound assumption while evaluating text quality of MT output. Kalyani and Sajja (2015) reviewed Indian MT systems and different evaluation methodologies [7]. They gave detailed insights about MT approaches like RBMT, EBMT, SMT and Hybrid approach. They also highlighted human and automatic evaluation methods in use. Khan et al. (2017) analysed various phrase-based SMTs on multiple Indian languages in their survey to promote the development of SMT and linguistic resources [8]. They highlighted that lack of data resources was the main reason of flaws in the MT systems of that time. Large parallel corpora were necessary to reliably estimate probabilities of different translations.

Revannuru et al. (2017) highlighted the use of neural MT to overcome the flaws of SMT systems. They created NMT for 6 Indian language pairs [9]. They compared evaluation results from METEOR and BLEU as well as UNK count and F-measure. They demonstrated good accuracy by comparing and outperforming Google translate by a margin of 17 BLEU points. Modh and Saini (2018) studied different approaches for machine translation in Gujarati language [10]. Their study included rule-based, empirical-based and hybrid MT approaches. They compared several MT systems like MANTRA, Google Translate, ANUVADAKSH and Angla Bharati (version I and II) on the basis of approach, language-pair, data quality and features.

Shah and Bakrola (2019) presented Neural Machine Translation System that can effectively translate Indian languages involving Hindi and Gujarati [11]. The system's performance was evaluated using automatic assessment metrics such as BLEU, Perplexity and TER. In terms of English-Gujarati translation, the developed system surpassed Google Translate by a margin of 6 BLEU scores. The absence of an index for evaluating language quality without reference has hindered the progress of Nearest Neighbors (NLG) models. In response to this issue, Ethayarajh and Sadigh (2020), suggested a model named as BLEU Neighbors, that includes BLEU score into sentence space as a kernal function [12]. The developed model had better results than human annotations when predicting the quality of conversation and automatically assessing essays.

Gandhi et al. (2021) examined several machine translation approaches and their use in various translation systems [13]. The approaches were compared based on characteristics like machine translation techniques, language combination and paper features. The study introduced a language translation model that utilised hybrid machine translation approach, to facilitate the process of translating English from Gujarati. Chauhan and Daniel (2023) conducted an extensive review on completely automated assessment metrics used in assessing machine generated outputs [14]. The metrics were classified into five different categories: semantic, syntactic, character, lexical and semantic & syntactic. The research also tackled difficulties in assessing machine generated outputs using statistical machine translation and neural machine translation. Also discussed the positive aspects, drawbacks, and limitations of metrics.

Joshi and Katyayan (2023) introduced an MT evaluation metric based on contextual embeddings and linguistic knowledge [15]. The performance of developed metric was evaluated using variety of language pairs like English-Hindi, English-Urdu, English-Gujarati, English-Marathi, and English-Odia. However, the developed evaluation metric had performed better than human evaluation across all the language pairs mentioned above. Focused on the rising demand for research in the field of multilingual translation, Sani et al. (2024) provided an extensive analysis of machine translation techniques for Indian languages [16]. The study provides an in-depth overview of the issues encountered by researchers. Also discussed several metrics used in MT to develop an exhaustive assessment system.

## Experimental Setup

For developing a supervised learning-based machine translation evaluation metric, we needed data on which the model could be trained. For this, it was required to get MT outputs from a variety of MT systems. This was achieved by capturing MT outputs from several MT systems viz Google Translate, Bing Translate, Angla Bharti MT system (developed by a consortium headed by IIT Kanpur), Anuvadaksh MT system (developed by a consortium headed by C-DAC Pune), Matra MT system (developed at C-DAC Mumbai), Sata Anuvadak (developed at IIT Bombay), EBMT system (developed at Banasthali Vidyapith). Among these, for some MT engines, the same set of MT outputs were captured in different time intervals. For Google's MT system, MT outputs were captured in February 2010 then again in January 2014 and then in January 2024. Similarly for Bing Translate, outputs were registered in January 2014 and January 2024. For the Anuvadaksh MT system, outputs were captured in April 2012 and January 2014.

Besides these, some MT systems were also trained using open-source toolkits viz MOSES and OpenNMT. For MOSES; phrase-based MT system, hierarchical phrase-based MT system and syntax-augmented MT system was trained. For OpenNMT a seq2seq MT system and a transformer-based MT system were trained. They all were trained for English-Gujarati language pair. Thus, we had 16 MT systems which 5000 English sentences and created a corpus of 80000 system-generated translations. This corpus consisted of 1500 sentences from the administration domain, which were taken from Rajya Sabha proceedings, 1500 sentences from the health domain, and 2000 sentences from the tourism domain, which were taken from the EILMT corpus. These translations were provided to human annotators for performing manual evaluations. We used HEval measure for this study which used 11 parameters for identifying the quality of MT system output. These parameters are shown in table 1.

| S.No. | Parameters |
|---|---|
| 1 | Translation of Gender and Number of the Noun(s). |
| 2 | Translation of tense in the sentence. |
| 3 | Translation of voice in the sentence. |
| 4 | Identification of the Proper Noun(s). |
| 5 | Use of Adjectives and Adverbs corresponding to the Nouns and Verbs. |
| 6 | Selection of proper words/synonyms (Lexical Choice). |
| 7 | Sequence of phrases and clauses in the translation. |
| 8 | Use of Punctuation Marks in the translation. |
| 9 | Fluency of translated text and translator's proficiency. |
| 10 | Maintaining the semantics of the source sentence in the translation. |
| 11 | Evaluating the translation of the source sentence (With respect to syntax and intended meaning). |

Table 1. HEval Measure's Evaluation Parameters

A Likert scale of 0-4 was used to judge these parameters. Finally, all these 11 parameters were averaged to get one final objective score. The 5000 English sentences were also given to human experts for manual translations. These were the reference translations with which the MT outputs were compared. Thus, we developed a dataset of 80000 system-generated outputs with their human evaluation score and reference translations and English source sentences.

## Proposed Methodology

To develop an automatic evaluation metric, machine-generated outputs were compared with reference translations and some features were extracted based on similarities. The list of features is shown in table 2.

| S.No. | Feature | Description |
|---|---|---|
| 1 | Lexical Cosine Similarity | Words of reference translation and MT output were compared. |
| 2 | POS Cosine Similarity | POS similarity was computed to examine the parts of speech similarity between reference and machine-generated outputs |
| 3 | Stem Cosine Similarity | The words in both candidate and reference translations were reduced to stems and their cosine similarity was also captured. |
| 4 | Word2Vec Cosine Similarity | Words both reference and candidate translations were converted into vectors compared. This captured the semantically similar words. |
| 5 | Sentence Embedding Cosine Similarity | The complete sentences were converted into embeddings using a pre-trained transformer model. |
| 6 | Statistical Language Model Probability | The candidate translation's LM probability was registered. |
| 7 | REKHA-1 Score | Evaluation Score of Automatic Evaluation metric which compared sentences for lexical, syntactic and semantic matching. |
| 8 | REKHA-2 Score | Evaluation Score of REKHA-1 coupled with the contextual embedding-based similarity |
| 9 | BLEU Score | Evaluation Score of BLEU metric |
| 10 | # content words reference | No. of content words in reference translation |
| 11 | # content words candidate | No. of content words in candidate translation |
| 12 | # content words source | No. of content words in the source sentence |
| 13 | Probability Unigrams in Q1 | Low-frequency single words |
| 14 | Probability Bigrams in Q1 | Combination of low frequency two words |
| 15 | Probability Trigram Q1 | Combination of low frequency of three words |
| 16 | Probability Unigrams in Q2 | Lower-medium frequency single words |
| 17 | Probability Bigrams in Q2 | Combination of lower-medium frequency two words |
| 18 | Probability Trigram Q2 | Combination of lower-medium frequency of three words |
| 19 | Probability Unigrams in Q3 | Medium-frequency single words |
| 20 | Probability Bigrams in Q3 | Combination of medium frequency two words |
| 21 | Probability Trigram Q3 | Combination of medium frequency of three words |
| 22 | Probability Unigrams in Q4 | High-frequency single words |
| 23 | Probability Bigrams in Q4 | Combination of high frequency two words |
| 24 | Probability Trigram Q4 | Combination of high frequency of three words |

Table 2. Feature set used for training of MaTrA

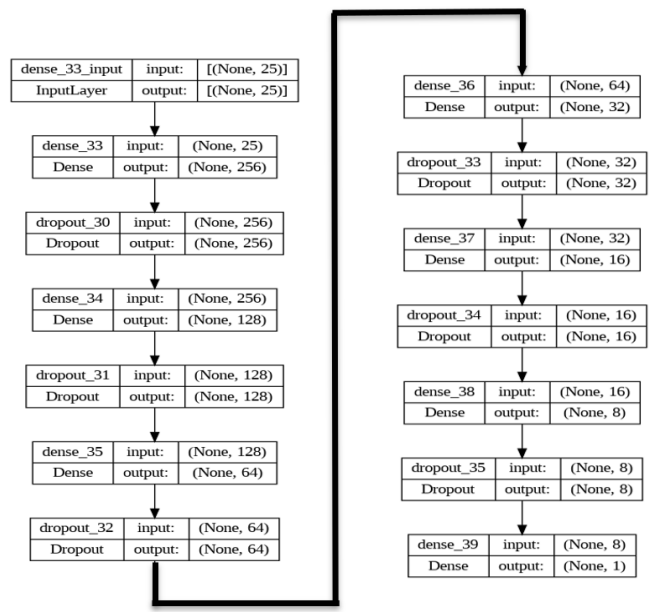

Figure 1: Model Summary of MaTrA-1

Figure 2: Model Summary of MaTrA-2

A dataset based on these 24 features and similarity scores was prepared which also the HEval average score as the target variable. Further, this dataset was used for training the model. To utilize the training data, we trained two deep-learning models. Among the two models, one model was trained using a deep neural network consisting 6 hidden layers, dropouts, L1 regularization, Batch Normalization, tanh activation function and Adam optimizer, calculating loss through MSE with 500 epochs while the other model was trained using an deep neural network consisting 10 hidden layers. The rest of the hyperparameters were the same as that of the previous variant. Figure 1 shows the model summary of MaTrA variant one while figure 2 shows the model summary of MaTrA variant two. The model architecture of these are shown in figure 3 and 4 respectively.

Figure 3: Architecture of MaTra-1

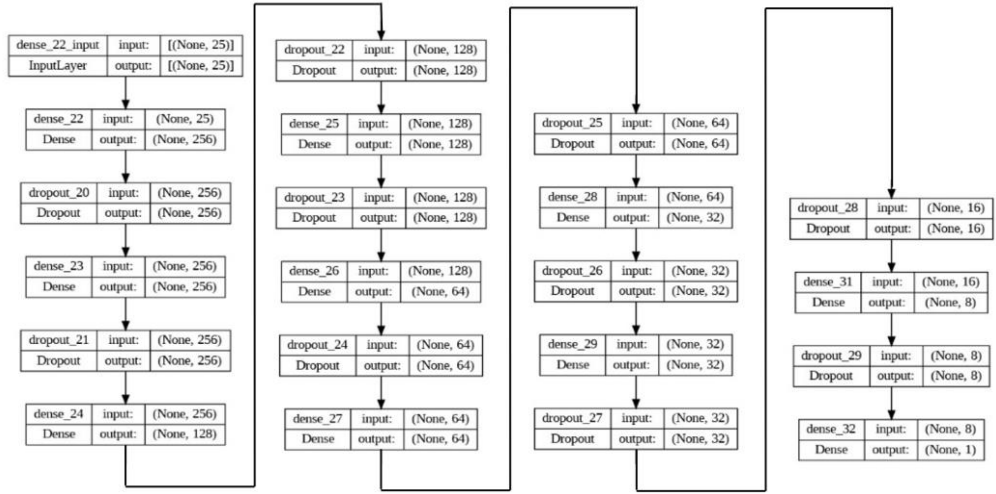

Figure 4: Architecture of MaTrA-2

## Evaluation

To understand the performance of the metric, we created a new corpus of 500 sentences each from the agriculture and education domains. We asked human annotators to provide reference translations for these 1000 sentences. Next, outputs from seven machine translation engines were registered. These MT systems were Google Translate, Bing Translate, Devnagri MT system, HimangY MT System, Digital India Bhashin Division's Anuvaad MT system, Chat GPT and AI4Bharat's MT system.

These MT system outputs were adjudged by popular MT evaluation metrics viz BLEU, Meteor, LEPOR, chrF++, COMET along with both the variants of MaTrA. We also performed a human evaluation of the MT system outputs using the HEval measure. Further, the results of MaTrA were also correlated with the HEval measure. Table 3 shows the comparative results of MaTrA with other MT evaluation metrics which are averaged at the system level. Table 4 shows the correlation of the Human evaluation measure (HEval) with MaTrA. In all the cases MaTrA-2 and MaTrA-1 showed better results as compared to other metrics. While correlating the results of the HEval with MaTrA-1 and MaTrA-2 also shows a positive correlation which suggests that these metrics can produce results comparable to human judgments.

| MT System | BLEU | Meteor | LEPOR | chrF++ | COMET | MaTrA-1 | MaTrA-2 |
|---|---|---|---|---|---|---|---|
| Google Translate | 0.002791 | 0.488718 | 0.759978 | 0.544282 | 0.801165 | 0.895418 | **0.931004** |
| Bing Translate | 0.002762 | 0.499412 | 0.777054 | 0.579330 | 0.821037 | 0.905367 | **0.941591** |
| Devnagri MT System | 0.002761 | 0.455151 | 0.771461 | 0.552161 | 0.810725 | 0.901543 | **0.944056** |
| HimangY MT System | 0.002767 | 0.483536 | 0.781413 | 0.548347 | 0.830689 | 0.902354 | **0.942564** |
| DIBD Anuvad MT System | 0.002789 | 0.532439 | 0.795129 | 0.573188 | 0.841644 | 0.900407 | **0.941311** |
| Chat GPT | 0.002721 | 0.279149 | 0.666296 | 0.405811 | 0.749840 | 0.903391 | **0.921810** |
| AI4Bharat MT System | 0.002789 | 0.532439 | 0.795129 | 0.573189 | 0.841644 | 0.900407 | **0.941311** |

Table 3. System Level Scores

| MT System | HEval-MaTrA-1 | HEval-MaTrA-2 |
|---|---|---|
| Google Translate | 0.755374 | 0.958647 |
| Bing Translate | 0.742807 | 0.940660 |
| Devnagri MT System | 0.435701 | 0.503484 |
| HimangY MT System | 0.589666 | 0.778742 |

| MT System | HEval-MaTrA-1 | HEval-MaTrA-2 |
|---|---|---|
| DIBD Anuvad MT System | 0.609103 | 0.693858 |
| Chat GPT | 0.487654 | 0.423543 |
| AI4Bharat MT System | 0.609103 | 0.693858 |

Table 4. Correlation with Human Judgement

## Conclusion

In this paper, we have shown the development of a reference-based machine translation evaluation metric using supervised learning. For the development of this metric 80000 sentences were translated across MT systems. The translations were then manually evaluated by human annotators based on 11 parameters and their average score was registered. Further human annotators were also asked to provide human reference translations for the source English sentences. The candidate translations and the human references were then used to extract features. 24 features were extracted which were used as input for the deep learning model and the averaged human evaluation was considered as the target. Two DNN models were trained, one had six hidden layers and the second had 10 hidden layers.

For testing of the trained models, 1000 sentences were identified from the agriculture and education domains. Seven MT engines were identified, and their outputs were captured. Human annotators were asked to provide reference translations for these 1000 sentences. Once all these were ready, five popular MT evaluation metrics were used along with the two trained DNN models for generating the evaluation scores. It was found that the trained DNN models performed better than the available MT evaluation metrics. To further strengthen this claim the results of the DNN models were correlated with human evaluation measures. It was found that the trained models produced comparable results with human judgments as in all the cases they showed a positive correlation.

In the future, we would like to extend these trained metrics to other Indian languages as well. We also wish to experiment with more DNN models to perfect our approach. For this, we would be using different sets of optimizers, different combinations of regularizes and different dropout values.

## Acknowledgment

This work is supported by the funding received from the Ministry of Electronics and Information Technology, Government of India for the project "English to Indian Languages and vice versa Machine Translation System" under National Language Translation Mission (NLTM): Bhashini through administrative approval no. 11(1)/2022-HCC(TDIL) Part 5.

## References

1. Gohil, L., & Patel, D. (2019). A sentiment analysis of gujarati text using gujarati senti word net. International Journal of Innovative Technology and Exploring Engineering (IJITEE), 8(9), 2290-2293.
2. Papineni, K., Roukos, S., Ward, T., & Zhu, W. J. (2002, July). Bleu: a method for automatic evaluation of machine translation. In Proceedings of the 40th annual meeting of the Association for Computational Linguistics (pp. 311-318).
3. Lin, C. Y., & Och, F. J. (2004). Orange: a method for evaluating automatic evaluation metrics for machine translation. In COLING 2004: Proceedings of the 20th International Conference on Computational Linguistics (pp. 501-507).

4. Banerjee, S., & Lavie, A. (2005, June). METEOR: An automatic metric for MT evaluation with improved correlation with human judgments. In Proceedings of the acl workshop on intrinsic and extrinsic evaluation measures for machine translation and/or summarization (pp. 65-72).
5. Gupta, A., Venkatapathy, S., & Sangal, R. (2010). METEOR-Hindi: automatic MT evaluation metric for hindi as a target. In Proceedings of ICON-2010: 8th international conference on natural language processing, Macmillan Publishers. India.
6. Joshi, N., Mathur, I., Darbari, H., & Kumar, A. (2013). HEval: Yet another human evaluation metric. arXiv preprint arXiv:1311.3961.
7. Kalyani, A., & Sajja, P. S. (2015). A review of machine translation systems in india and different translation evaluation methodologies. International Journal of Computer Applications, 121(23).
8. Khan, N. J., Anwar, W., & Durrani, N. (2017). Machine translation approaches and survey for Indian languages. arXiv preprint arXiv:1701.04290.
9. Revanuru, K., Turlapaty, K., & Rao, S. (2017, November). Neural machine translation of indian languages. In Proceedings of the 10th annual ACM India compute conference (pp. 11-20).
10. Modh, J. C., & Saini, J. R. (2018). A STUDY OF MACHINE TRANSLATION APPROACHES FOR GUJARATI. International Journal of Advanced Research in Computer Science, 9(1).
11. Shah, P., & Bakrola, V. (2019, February). Neural machine translation system of indic languages-an attention based approach. In 2019 Second International Conference on Advanced Computational and Communication Paradigms (ICACCP) (pp. 1-5). IEEE.
12. Ethayarajh, K., & Sadigh, D. (2020). Bleu neighbors: A reference-less approach to automatic evaluation. arXiv preprint arXiv:2004.12726.
13. Gandhi, V. A., Gandhi, V. B., Gala, D. V., & Tawde, P. (2021, October). A study of machine translation approaches for gujarati to english translation. In 2021 Smart Technologies, Communication and Robotics (STCR) (pp. 1-5). IEEE.
14. Chauhan, S., & Daniel, P. (2023). A comprehensive survey on various fully automatic machine translation evaluation metrics. Neural Processing Letters, 55(9), 12663-12717.
15. Joshi, N., & Katyayan, P. (2023, May). Rekha: A Reference Based Machine Translation Evaluation Metric Using Linguistic Knowledge and Contextual Embeddings. In ICIDSSD 2022: Proceedings of the 3rd International Conference on ICT for Digital, Smart, and Sustainable Development, ICIDSSD 2022, 24-25 March 2022, New Delhi, India (p. 158). European Alliance for Innovation.
16. Sani, S., Vijaya, S., & V Gangashetty, S. (2024). A Survey on the Machine Translation Methods for Indian Languages: Challenges, Availability, and Production of Parallel Corpora, Government Policies and Research Directions. International Journal of Computing and Digital Systems, 15(1), 1-11.